\documentclass[a4paper,11pt]{article}

\usepackage{fancyhdr}
\usepackage{latexsym}
\usepackage{amssymb}
\usepackage{fourier}
\usepackage{color}
 \usepackage{graphicx}
\usepackage{url}
\usepackage[affil-it]{authblk}
\usepackage{amsmath}
\usepackage{wrapfig}

\usepackage[T1]{fontenc}
\usepackage{times}

\usepackage{url}
\usepackage{xcolor}
\definecolor{newcolor}{rgb}{.8,.349,.1}

\begin{document}

\title{ A spatial hue similarity measure for assessment of colourisation}

\author[1,2]{Se\'an Mullery}
\author[1]{Paul F. Whelan}
\affil[1]{Vision Systems Group, School of Electronic Engineering, Dublin City University, Dublin 9, Ireland}
\affil[2]{IT Sligo, Ash Lane, Sligo,
Ireland}

\date{}

\maketitle
\begin{center}
    \textit{This paper is under consideration for Pattern Recognition Letters.}
\end{center}

\begin{abstract}
Automatic colourisation of grey-scale images is an ill-posed multi-modal problem. Where full-reference images exist, objective performance measures rely on pixel-difference techniques such as MSE and PSNR. These measures penalise any plausible modes other than the reference ground-truth; They often fail to adequately penalise implausible modes if they are close in pixel distance to the ground-truth; As these are pixel-difference methods they cannot assess spatial coherency.
We use the polar form of the a*b* channels from the CIEL*a*b* colour space to separate the multi-modal problems, which we confine to the hue channel, and the common-mode which applies to the chroma channel. We apply SSIM to the chroma channel but reformulate SSIM for the hue channel to a measure we call the Spatial Hue Similarity Measure (SHSM).  This reformulation allows spatially-coherent hue channels to achieve a high score while penalising spatially-incoherent modes. This method allows qualitative and quantitative performance comparison of SOTA colourisation methods and reduces reliance on subjective human visual inspection.  
\end{abstract}


%\linenumbers
\renewcommand{\thefootnote}{\roman{footnote}}
\section{Introduction}
\label{sec1}
Colourisation is the process of inferring colour information from a grey-scale prior and then combining the colour, with the prior, to produce a full-colour image. It is an ill-posed problem. For every prior, there are many plausible colourisations and many more that are implausible. Manual colourisation, as carried out by humans, relies on heuristics, common sense and prior knowledge. There is a large body of work on semi-automatic and automatic colourisation. The most recent work in colourisation involves the use of Generative Adversarial Networks (GANs) \cite{goodfellowGAN2014}, in which one deep neural network ($\mathcal{G}$) tries to create plausible colourisations and a second network ($\mathcal{D}$) tries to discriminate between real colourisations and those produced by $\mathcal{G}$. For the colourisation task, GANs try to learn the manifold on which all the plausible colourisations lie. For training, a set of natural images is converted to a colour space that separates the grey-scale from the colour information; CIEL*a*b* is the most popular for this. $\mathcal{G}$ is then trained to produce a*b* from an L* prior. This training regime, only admits a single plausible colourisation for each prior, which is limiting, when trying to learn the full manifold of all plausible colourisations. There have been some attempts to measure the performance of colourisations, but all are problematic, particularly when it comes to applying a similar score to multiple plausible colourisations. This paper will present one possible measurement system.

\section{Measuring Colour}
Most colourisation techniques rely on some form of human visual inspection to determine efficacy, or for comparison to other techniques, \cite{IizukaLetThereBeColor:2016:LCJ:2897824.2925974}, \cite{IronyColorByExample2005}, \cite{Isola2017}, \cite{WelshTransferingColor:2002:TCG:566654.566576},  \cite{YatzivChrominanceBlending1621234},  \cite{ZhangIE16ColorfulImageColorization}, \cite{ZhangZIGLYE17UserGuidedColorization}.  Human visual inspection can include qualitative analysis, naturalness scoring, or real/fake judged by human. In some cases the Amazon Mechanical Turk (AMT) was used with up to 40 participants looking at each technique, \cite{ZhangIE16ColorfulImageColorization}.

%Those that used scribble or hint inputs of the ground-truth colour \cite{LevinColorizationOptimization2004:CUO:1015706.1015780}

Many attempt an objective measure based on pixel value errors, such as RMS pixel error, \cite{Deshpande7410429LargeScaleAutoColorization}, PSNR \cite{Cheng2015DeepC},  \cite{ZhangZIGLYE17UserGuidedColorization}, L2 pixel distance  \cite{ZhangIE16ColorfulImageColorization} (only within a defined threshold).  Some methods \cite{ZhangIE16ColorfulImageColorization}, \cite{larsson2016learning}, \cite{Vitoria_2020_WACV}  utilise the concept that colour assists in classifying objects. Therefore a network designed to classify objects using colour images will show a deterioration in performance if inferred with a poorly colourised image. The difference can then be used as a proxy measure for colourisation performance.
From the results of \cite{ZhangIE16ColorfulImageColorization}, it is clear that while correct colour images achieve higher classification scores than their equivalent grey-scale image, the grey-scale image still makes up the majority of the information required for classification. E.g. grey-scale 52.7\% vs ground-truth colour 68.3\%. 

As GANs are now a primary focus of colourisation research, it is worth mentioning two measures, Fr\'echet Inception Distance, \cite{HeuselRUNKH17} and Inception Score, \cite{SalimansNIPS2016_6125}, that are starting to gain a consensus in GAN literature. However, it is clear that they are imperfect measures and have not, to our knowledge, been used for the judgement of the fidelity of colourisations.

\subsection{Problems in measuring colourisations}
The first problem with measuring colourisation is that it is a multi-modal problem. While those colourisation methods that employ hints, from the user, require a single colour target for each grey-scale prior, fully-automatic colourisation allows for many plausible colourisation targets.

The second problem is that of spatial coherency. The colour should remain consistent within a given semantic region, and then change precisely at the boundary between regions.
%This is analogous to measuring the performance of segmentation algorithms where the Intersection over Union (IoU AKA Jaccard index) or Boundary Jaccard, \cite{boundaryJaccard}, may be used to score the segmentation algorithm. Segmentation, unlike colourisation,  has a single target and are also uniform, whereas a colour region can contain texture which a colourisation algorithm would be expected to reproduce.

Pixel difference methods (L1, L2, PSNR, MSE, RMS), measure only the difference at a pixel, in isolation from its neighbours, between a colourisation and a single ground-truth. So they penalise any sample from the distribution of plausible colourisations that is not the ground-truth . They fail to penalise for poor spatial coherency as they don't take account of neighbouring pixels, \cite{MulleryWhelan2019}.
This has led to the wide-spread use of naturalness measures based on human inspection.

A third problem, which we do not consider in this work, is the plausibility of the modes of colour for a given semantic object, e.g. Blue and Green are implausible colours for skin tones, but plausible for clothing. 

\subsection{What makes a plausible colourisation?}
As stated above, we will not look at the plausibility of colour modes for specific semantic objects. The section on future work will outline potential routes forward which may benefit from the work in this paper. Instead, we will make the simplifying assumption that plausible colourisations can take on any hue, as long as that hue is consistent within the borders of the semantic object.
We define consistent as maintaining the same small value changes (texture) in hue as the ground-truth within the borders of a semantic segment and changing the hue value by a significant amount at the same spatial locations as these changes in a ground-truth image.

The saturation/chrominance of an object depends on the object itself, its reflectance properties (specular/lambertian) and the lighting conditions. Human colourists consider all of these things when deciding what level of saturation to apply. With this in mind, we can see that each object has a narrower distribution of plausible saturations that could be applied, compared to hue. We make the simplifying assumption that the saturation of an object should be the same in all colourisations as in the ground-truth.
It is difficult to make such simplifying assumptions about the CIEL*a*b* space, as equal chrominance (a close relative of saturation) forms a circular path around the centre of a*b* space.
However, CIEL*a*b* attempts to provide a coordinate system between two stimuli that is close to the colour difference perceived by the human eye over small distances, \cite{BRAINARD2003191}, and therefore we will convert CIEL*a*b* to $L^*h_{ab}^{*}c_{ab}^{*}$, \cite{fairchild.ch10}, using the following equations.
\begin{equation}
c_{a b}^{*}=\sqrt{\left(a^{* 2}+b^{*} 2\right)}
\end{equation}
\begin{equation}
h_{a b}^{*}=\tan ^{-1}\left(b^{*} / a^{*}\right)
\label{hab}
\end{equation}
For the rest of this paper, $h_{a b}^{*}$ will be referred to as hue, and $c_{a b}^{*}$ will be referred to as chroma.

\section{Proposed method}
We propose to measure the chroma using the Structural Similarity Index Measurement (SSIM), \cite{SSIM}. SSIM is designed to measure the perceived quality. It takes into account local spatial comparisons of magnitude/luminosity, contrast and structure. We give each of these equal precedence, as did the original authors of SSIM (see equation (\ref{SSIMComb}) below, $\alpha=\beta=\gamma=1$). 
%With further research, we may find more nuanced values for $\alpha, \beta$ and $\gamma$ that better represent the distribution of chroma for a given prior.
\begin{equation}
SSIM(x, y)=[l(x, y)]^{\alpha} \cdot[c(x, y)]^{\beta} \cdot[s(x, y)]^{\gamma}
\label{SSIMComb}
\end{equation}

We cannot use SSIM to measure hue for the following reasons. Firstly, the magnitude/luminosity measure of SSIM will penalise in the same way as the pixel difference measures, and this means those plausible hues that are not close to the ground-truth will be penalised. 
Secondly, the local structure measurement is also problematic in that it requires the sign of local changes to be identical between the images being compared. Local changes in hue, for two plausible colourisations, would not need to be of the same sign.

The contrast measurement of SSIM is more promising, as two plausible colourisations might be expected to have the same contrast in hue within a semantic segment of the image. A problem occurs at the border of two segments. The contrast here will be relatively large, but the actual magnitude of the difference would not necessarily be the same between two plausible colourisations. So our third issue, with SSIM for hue, is that it cannot treat all large changes in hue as being similar, which is necessary to compare plausible colourisations.

\subsection{Other nuisances of hue space}
Hue represents an angle. There is no hard edge in the CIEL*a*b* colour space. Hue here should not be mistaken with the idea of a specific frequency of colour. Colour frequencies do not rotate back circularly. However, from the perspective of human perception, a combination of Red and Blue (opposite ends of the spectrum), is possible and as such the hue does complete a continuous periodic space. If two hues are quite close together and so may represent a colour texture, they may be represented by numbers that are far apart. Using equation (\ref{hab}) we will obtain hue values $\in [-\pi,\pi]$. However, in our space, $-\pi$ and $\pi$ are zero distance apart. For any number system we choose, these will be considered the maximum distance apart. We instead, ensure that the shorter distance around the circle is always chosen with what we will refer to as an \textit{Angle\_Diff} function, see equation (\ref{angleDiff}). 

A second problem to consider is the hue of negligible chroma values. In circumstances where the chroma value is minimal, then the specific hue has little meaning. Even large changes in hue will not be noticeable to a viewer if the associated chroma at that location is very small. Despite this, we have no mechanism to convey an absence of hue. In many cases, the hue can be noisy and have changes that appear significant when analysed without consideration of the chroma. This occurs when negligible values of a* and b* can be modified by small rounding errors in calculation, often leading to large changes in hue using equation (\ref{hab}). There is no added utility for a colourisation algorithm to attempt to generate these invisible changes in hue. In our proposed algorithm, we will multiply the gradients of the hue channel by a logistic function of the chroma value for each pixel location to disregard gradients in the hue that corresponds to negligible chroma values.

\subsection{Spatial Hue Similarity Measure SHSM}
\begin{equation}
	\nabla h'= \mbox{\textit{Angle\_Diff}}\left(\left|\nabla  h\left(\frac{1}{1+e^{-k\left(c-c_{0}\right)}}\right)\right|\right)
	\label{angleDiff}
\end{equation}

$h$ is the hue channel, and $c$ is the chroma channel. $c_0$ is the user selected centre of the chroma range that we use for the logistic function to reduce sensitivity to hues that correspond to minimal chroma values. $k$ is the slope of the curve at its mid-point. %We chose a value of 5 in this work. 
As hue is a circular space, the \textit{Angle\_Diff} function is used to find the smallest gradient that represents the change in hue. We use the absolute value of hue change $(| |)$ to be insensitive to the direction of change in the hue space.

\begin{equation}
\nabla h_n=\frac{1}{1+e^{-k\left(\nabla h'-h_{0}\right)}}
\label{logistic}
\end{equation}

Here we transform the $\nabla h'$ through another logistic function. $h_{0}$ is the chosen mid-point of the curve and $k$ is the slope of the curve at its mid-point. The output of this function will be $\mathbb{R} \in [0,1]$. We denote this, $\nabla h_{n}$ as it is the hue gradients normalised to the range $[0,1]$.

Transformation, through the logistic function, attenuates the significance of the smallest hue changes that are less perceptible to the human eye. It also extends the significance of small to medium changes in the hue that would represent perceptually visible hue texture. Finally it constrains, to a value of $1$, all hue changes in the medium-to-large range that would be consistent with a change in the semantic region. 

Just as in SSIM, to determine spatial similarity we will convolve ($\circledast$) $\nabla h_{n}$ with a 2D Gaussian $g$, we retain the values for this Gaussian as used in the original SSIM default formulation, i.e. $\sigma=1.5$, \cite{SSIM}.
                
\begin{equation}
	\nabla h_{s} = g \circledast \nabla h_n
	\label{}
\end{equation} 
Where $\nabla h_{s}$ represents the gradient of spatially weighted hue.

Finally we use the contrast part of SSIM to do our hue comparison.
\begin{equation}
	\textit{SHSM}(GT, CL) = \frac{2 \nabla h_{s}(GT)\nabla h_{s}(CL) + \epsilon}{\nabla h_{s}(GT)+\nabla h_{s}(CL) + \epsilon}
	\label{hueContEq}
\end{equation}

Here, $GT$ is the ground-truth image, and $CL$ is the colourisation that is being compared. As in SSIM, $\epsilon$\footnote{In the original SSIM paper this was called C, but we do not want this to confuse our notation where c is chroma.} is included for numerical stability but can also be used to softly ignore small, less perceptible, differences between the two images. We choose a value of 0.03\footnote{In SSIM, this was the value for K and C was derived from that with C=L*K where L was 255 for an 8-bit image. However, as we have normalised all values to 1, then C and K are the same value}, which is what SSIM chooses for the contrast part of the calculation.
We must calculate the gradients in two orthogonal directions (we choose the two diagonals), and so we will calculate $\textrm{SHSM}$ for the two directions and multiply the results. 

\subsection{Hyper-parameter Choices}
\begin{figure}[!t]
\centering
\includegraphics[width=.5\textwidth]{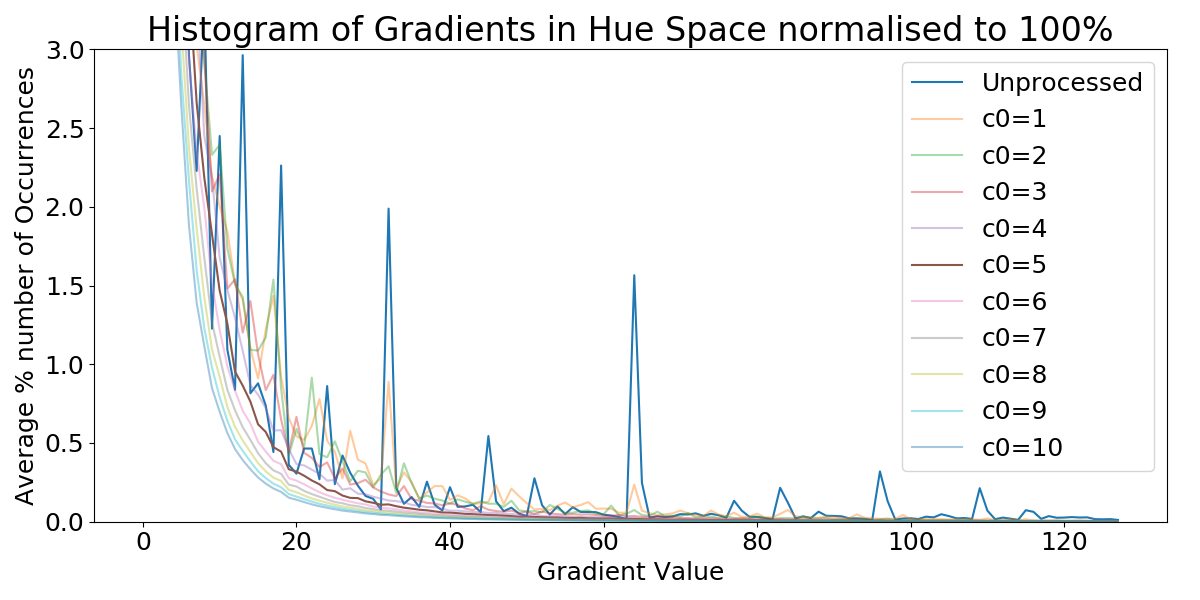}
\caption{Here we see the histogram of occurrences of gradients in hue space for various choices of $c_0$. In the unprocessed case, we see several spikes in the number of occurrences whose positions can all be explained by small values in the a*b* channels. We have highlighted $c_0=5$ as the first of the curves that seem to remove these spikes, but otherwise not further concentrate the number of occurrences in the smaller gradient values.}
\label{removeNuicance}
\end{figure}

To analyse the hue channel, 4000 natural images from the Places dataset, \cite{Places}, are converted to the $L^*h_{ab}^{*}c_{ab}^{*}$ space. The gradient of the hue channel is what is of interest, so a histogram of gradients in two orthogonal directions is constructed. 
We normalise the histogram to show the percentage of pixels on the y-axis, see Fig \ref{removeNuicance}. The graph labelled 'Unprocessed' has many large spikes throughout the range, showing that specific gradients are very prevalent throughout the dataset. Each of these can be explained by a specific case of changes in small values of chroma. e.g. when converting from CIEL*a*b*, if $(a=0,b=0)$ changes to $(a=0,b=1)$ or $(a=1,b=0)$ we get a hue of $|\pm\frac{\pi}{2}|$ which is 63 when represented with 8-bits covering $-\pi \textrm{ to} +\pi$. This case represents the peak visible in the centre or the distribution. Similar peaks occur for combinations of $a$ and $b$ for values $0-6$. These do not represent accurate hue values but are instead a nuisance of the conversion to a polar form and as the chroma is a small value the hue change will be difficult to perceive. To remove the influence of these, we pass the hue gradients through a logistic function, Equation (\ref{angleDiff}). To choose an appropriate value of $c_0$ we reprocess the dataset with values of $c_0 \in [1,10]$ and re-construct the histogram, see Fig \ref{removeNuicance}.
 Increasing values of $c_0$ smooth the histogram.
 %revealing an exponential decrease in the number of occurrences of gradients with increasing gradient magnitude.
We choose a value of $c_0=5$, as this appears to be the first of the lines to lose peaks in the distinctive locations of small chromas.

\begin{figure}[!t]
\centering
\includegraphics[width=.5\textwidth]{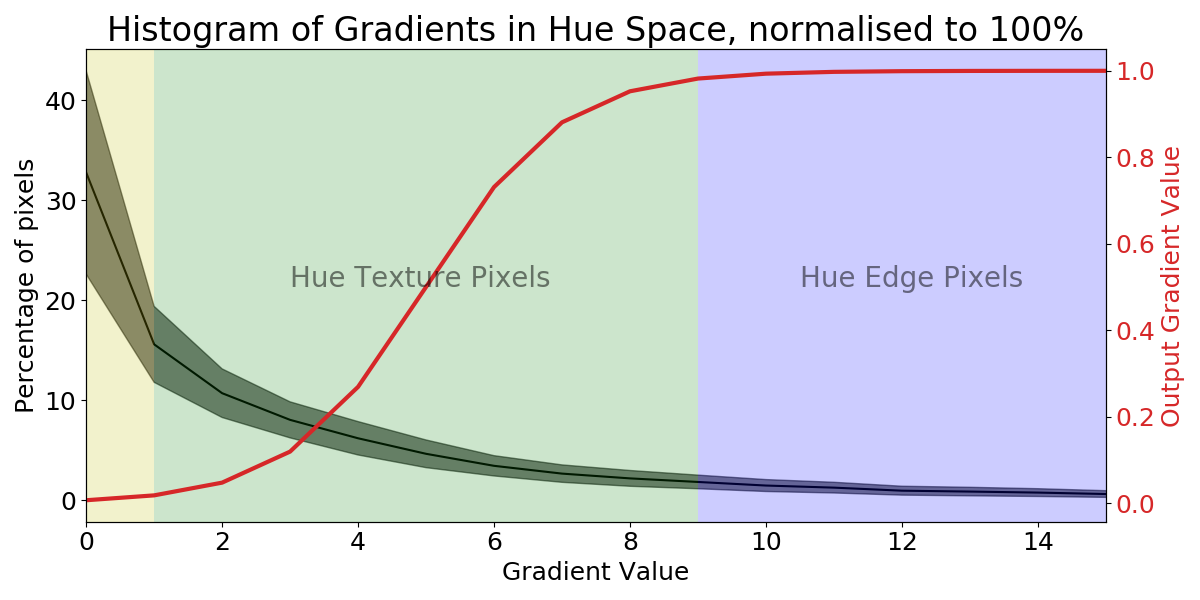}
\caption{Here, the black centre line of the curve represents the histogram of gradient values using $h_0=5$, and the grey area represents the standard deviation of the number of occurrences over the 4000 images. We have superimposed the logistic function, and the right-hand y-axis shows what input gradient values will become at the output of the logistic.}
\label{textureVBoundary}

\end{figure}

With these nuisances removed using a fixed value of $c_0=5$, we re-analyse the 4000 images to see how much variance is in the individual gradient histograms of the set, see Fig \ref{textureVBoundary}. 
This provides valuable insight into the statistics of the hue channel. The majority of pixels have a gradient of $\leq2$. Gradients of this size will be challenging for humans to perceive. This informs us that most of the hue channel consists of flat, featureless regions. There is a small, but a not-insignificant percentage of pixel gradients  $\in[2,9]$. 
This suggests that there are some regions of the hue channel that may represent a hue texture.
Gradients $>9$ appear to be rare in the data set with percentages of pixels $<1\%$.
Due to the decreasing percentage of pixels, it becomes increasingly less likely that these gradients represent a significant region of an image and more likely that they represent the edge of a region of hue. For the multitude of plausible hues, the gradient at these edges will vary significantly over an unknown distribution. For the SHSM measure, we want to treat all edge pixels as equally important. For this reason, we process the gradient values with a logistic function, see Equation (\ref{logistic}). With this, all values below $2$ will be treated as close to zero and values above $9$ will be close to $1$ and therefore all treated the same (see right-hand vertical axis Fig. \ref{textureVBoundary}).  Values in between will be given a sliding scale, which is close to linear, via the $k=1$ and a central value of $h_{0}=5$. These values will be assumed to be a texture that is visible to the viewer and therefore, should be similar in scale between the ground-truth  and a plausible colourisation.

\begin{figure*}[!t]
            \includegraphics[width=0.22\textwidth]{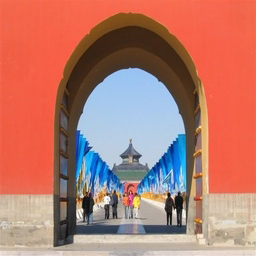}
           \includegraphics[width=0.22\textwidth]{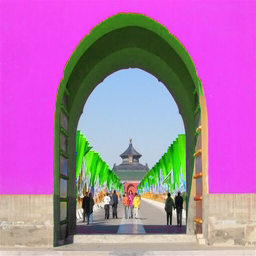}
            \includegraphics[width=0.22\textwidth]{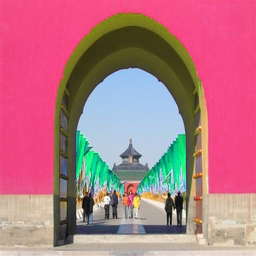}
            \includegraphics[width=0.22\textwidth]{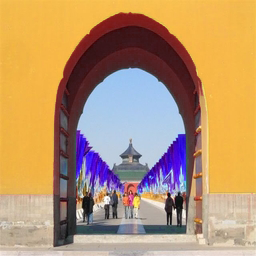}
            
            \hspace{0.225\textwidth}\includegraphics[width=0.22\textwidth]{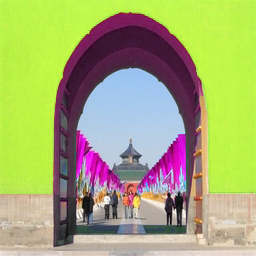}
            \includegraphics[width=0.22\textwidth]{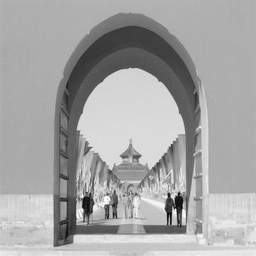}
            \includegraphics[width=0.22\textwidth]{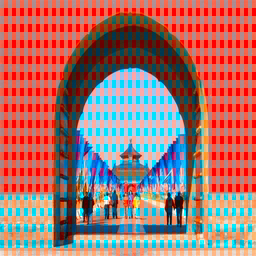}
    \caption{%
        A ground-truth  image (left) with four plausible colourisations produced in Adobe Photoshop\textsuperscript{TM} and two implausible colourisations.
     }%
    \label{psRecolour}
\end{figure*}

\begin{table*}[!t]
\begin{center}
 \begin{tabular}{|l|l|l|l|l|l|l|}
 \hline
 & \multicolumn{2}{c|}{MSE } &\multicolumn{2}{c|}{PSNR } & \multicolumn{2}{c|}{Proposed Method }\\
 \hline
                 & Hue & Chroma & Hue & Chroma & Hue & Chroma\\
                \hline
Colourisation 1 & 2610     &  394   & 13.96dB    &  22.17dB   & 0.902     & 0.915    \\
\hline
Colourisation 2 & 1660     &  22.41   & 15.9dB      &  34.6dB  & 0.944      &  0.962   \\
\hline
Colourisation 3 & 592     &  110   & 20.41dB     &  27.70dB   & 0.956      &  0.958   \\
\hline
Colourisation 4 &  2485     &   138 & 14.17dB     &  26.74dB   & 0.935      &  0.940   \\
\hline
Stripe Pattern      & 2588     &   728  & 14.00dB      &  19.51dB  & 0.484   & 0.208    \\
\hline
Grey-scale  & 2268     &  2152    & 14.57dB     &   14.8dB  & 0.499      &   0.033 \\
\hline
\end{tabular}
\caption{Four plausible Adobe Photoshop\textsuperscript{TM} colourisations and two implausible colourisations are compared to a ground-truth image. MSE/PSNR fail to separate plausible from implausible colouristions while the proposed method does.}
\label{psTableResults}
\end{center}

\end{table*}

\begin{figure*}[!t]
    \begin{center}
	\includegraphics[width=0.75\textwidth]{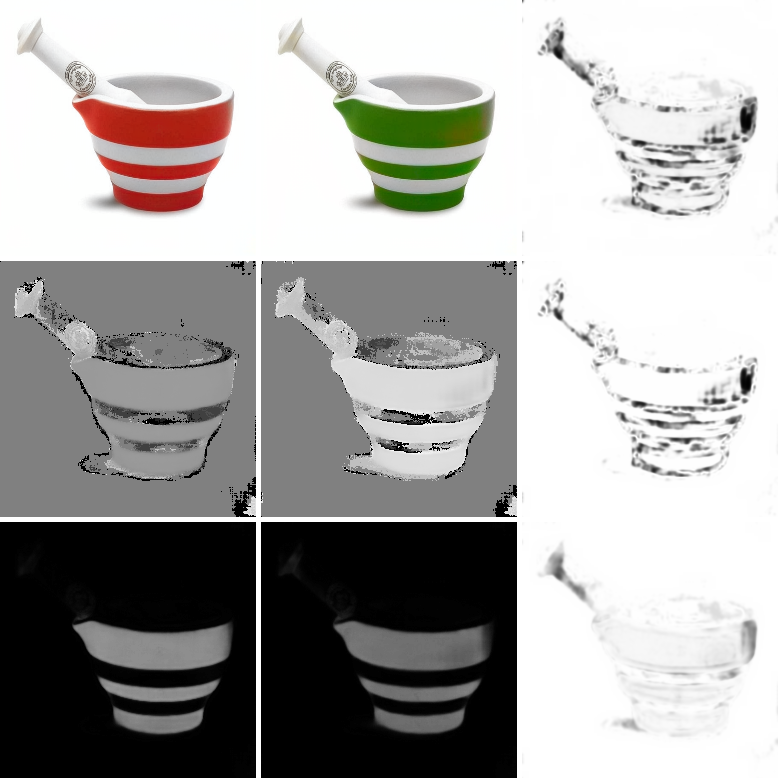}
	\end{center}
	\caption{The left-hand column is the first colourisation and the middle column the second colourisation, there is no ground-truth in this case. We have attempted, with user hints to make a red and green version. The third column shows the similarity using our method, where 0 (black) represents minimum similarity, and 255 (white) represents maximum similarity.  The hue has been penalised, particularly at the edges (middle row). We see that the chroma values produced are not the same (bottom row).}
	\label{mortorPestle}
 \end{figure*}

\begin{table*}[!t]
\begin{center}
\begin{tabular}{|l|l|l|l|l|l|l|}
\hline
 & \multicolumn{2}{c|}{MSE } &\multicolumn{2}{c|}{PSNR } & \multicolumn{2}{c|}{Proposed Method }\\
 \hline
 Mortor and Pestle image & Hue & Chroma & Hue & Chroma & Hue & Chroma\\
\hline
 Colourisation Comparison & 1634     &  27.21  & 16.0dB    &  33.78dB   & 0.963     &  0.978   \\
\hline

\end{tabular}
\caption{A comparison of two user-guided colourisations using the method of \cite{ZhangZIGLYE17UserGuidedColorization}, see Fig \ref{mortorPestle}}
\label{mortorPestleTableResults}
\end{center}

\end{table*}

\begin{figure*}[!t]
    
        \includegraphics[width=0.5\textwidth]{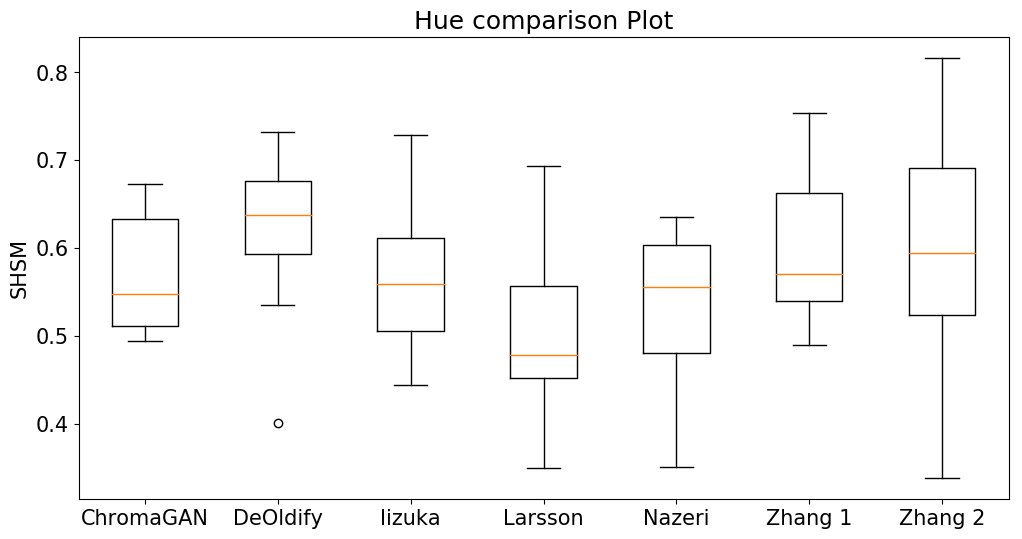}       
        \includegraphics[width=0.5\textwidth]{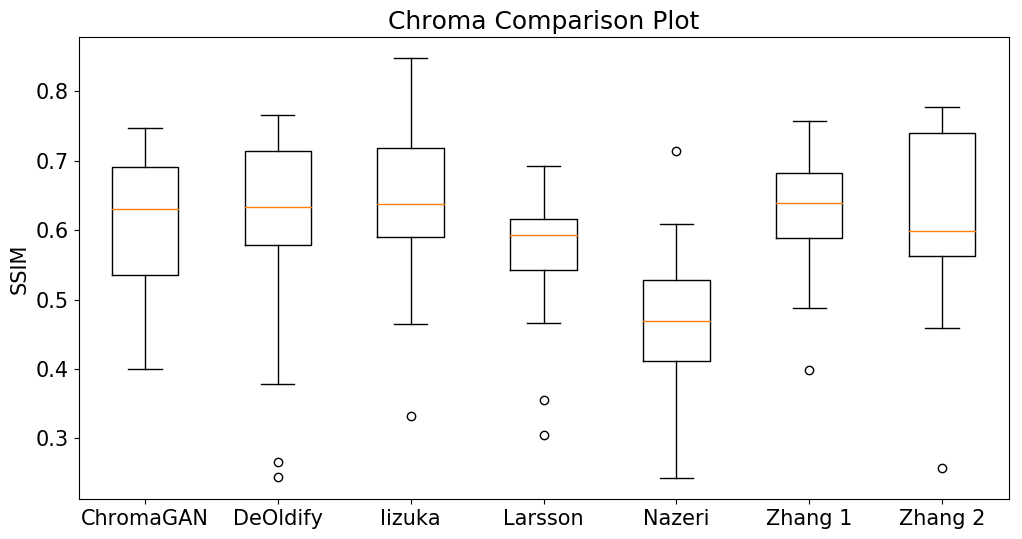}
        
	\caption{For 20 images from the Places dataset, we compare six methods' colourisations against the ground-truth .
	ChromaGAN, \cite{Vitoria_2020_WACV}, DeOldify\cite{DeOldify}, Iizuka,  \cite{IizukaLetThereBeColor:2016:LCJ:2897824.2925974}, Larsson, \cite{larsson2016learning}, Nazeri, \cite{nazeri2018image}, Zhang 1, \cite{ZhangIE16ColorfulImageColorization}, Zhang 2,  \cite{ZhangZIGLYE17UserGuidedColorization} (which is used without any user hints in this case).}
	\label{boxPlots}
	
 \end{figure*}

\section{Experimental Results}
Ideally, we would like to have multiple colour versions of each natural image to test the measurement system. Unfortunately, there is no ideal algorithm for colourisation or re-colourisation. Instead, we will firstly recolour specific sections of a ground-truth image in Adobe Photoshop\textsuperscript{TM} using the magic wand tool to select segments and then change the hue \footnote{Hue in Adobe Photoshop \textsuperscript{TM} will not be from the same space as $L^*h_{ab}^{*}c_{ab}^{*}$, so it will not only affect the hue in our $L^*h_{ab}^{*}c_{ab}^{*}$ experiments.}  of those segments. To the human eye, these look plausible, see Fig. \ref{psRecolour}. We will expect these to score well on similarity. We also include a grey-scale image, and an image with a stripe pattern, with each of the stripes being close to the original ground-truth colour. These look like poor colourisations to the human eye so should score poorly. Table \ref{psTableResults} shows a comparison of the score of each against the ground-truth on MSE, PSNR and our method for hue and chroma channels. 
For MSE, a lower number indicates more similarity. For PSNR and our method, the higher number suggests more similarity. While the change made in Adobe Photoshop\textsuperscript{TM} was only to hue, it also changed chroma. We see that both MSE and PSNR have quite a variability in chroma scores for the four colourisations, but they are better than the score for the stripe pattern or grey-scale as would be expected. With hue for MSE and PSNR, we see that some of the Adobe Photoshop\textsuperscript{TM} colourisations receive a worse or on-par score to the stripe and grey-scale.
For our method (standard SSIM) the chroma for the four colourisations has relatively small variance and is on the opposite end of the score range to the grey-scale and stripe pattern which our method has heavily penalised. For hue, we once again see broadly similar scores for the four plausible colourisations. The stripe pattern and grey-scale are much lower, but we note, not close to zero. This is because there are still many parts on the image where the gradient of hue is the same (i.e. small or zero), and this registers as similar. In general, we would advise taking the product of the two values (hue and chroma) as a total score of similarity of two colourisations.

In a second test, we do a qualitative analysis of what our algorithm penalises. We will use the work of  \cite{ZhangZIGLYE17UserGuidedColorization} with their user-guided colourisation, see Fig. \ref{mortorPestle}, to make two colourisations of a grey-scale image. There is a combination of user variability and non-determinism in the system that makes it difficult to produce two colourisations that are close to being as spatially consistent as the system may ultimately be capable of achieving. We want to use our system to see how similar it grades the two colourisations and to show where and why it penalises them. Note: we are comparing these against each other, but the order of the images in the comparison does not matter as the measure is symmetrical. 
 Fig. \ref{mortorPestle} shows
 that the chroma (middle row) is not the same for the green and red images and the algorithm has penalised areas in the white stripes in the hue comparison (bottom row), particularly at the borders. A close inspection of the green image shows that the green colour has bled into the white, which has not happened in the red image. Our algorithm has penalised for a section on the right of the top (larger) colour stripe. A close inspection shows a red tint in the green image, which has no correspondence in the red image. Despite this, the two images are very similar, and this does achieve a high score of 0.946 when chroma and hue results are combined by multiplication. Full results for the comparison of the two images in chroma and hue for MSE, PSNR and our method are shown in table \ref{mortorPestleTableResults}. Once again, MSE and PSNR would not rate these images as similar, but our method does.
 
 Our third test is to see how some of the recent deep-learning methods fare on our measure.
We use a small sample of 20 images\footnote{Some of the methods are only available via a web interface which doesn't allow for automated testing of large sets of images} from the Places dataset to test each method. We compare the models of \cite{DeOldify}, \cite{IizukaLetThereBeColor:2016:LCJ:2897824.2925974}, \cite{larsson2016learning}, \cite{nazeri2018image},\cite{Vitoria_2020_WACV},   \cite{ZhangIE16ColorfulImageColorization}, \cite{ZhangZIGLYE17UserGuidedColorization} (which is used without any user hints in this case).
The results for chroma (SSIM) and hue (SHSM) are shown in Fig. \ref{boxPlots}.

We can see that there is considerable variability in all methods, even over a small sample of 20 images. We should note that the results can be negatively affected by the compression quality of output from the system. DeOldify and Nazeri in particularly showed significant degradation of the L-channel when SSIM was used even though the L-channel should be identical. This is likely caused by compression in JPEG files, where some of the other methods allowed for png file output. The method of  \cite{ZhangIE16ColorfulImageColorization} and \cite{Vitoria_2020_WACV} also used JPEG but the file sizes are larger, suggesting less compression. JPEG compresses the colour information significantly more than the grey-scale information so we can expect that this negatively affects some of the methods' results.

\section{Future Work}
Our simplifying assumption that chroma would share a single-mode between all colourisations, while necessary for the present, is difficult to verify. We hope that through experimentation with this current measure, we can explore and reduce our uncertainty with the chroma channel.
	
The method we presented here does not attempt to measure whether the hue applied to a semantic segment is plausible for a natural image. This would require knowledge of all object types and the probability distributions of colour tones for each, and measuring the divergence of the predicted distributions from the plausible distributions. %Given the availability of deep learning models for classification, localisation, and semantic segmentation  that are trained on large data sets, we could employ these to produce hue distributions for the classified objects. 
ChromaGAN, \cite{Vitoria_2020_WACV}, attempted to learn this distribution implicitly by minimising the KL divergence of the class distribution with a pre-trained VGG16 network, but this could be attempted explicitly with the hue distribution for classes. 
	
%Our method only considers spatial coherence over one scale. Further work could consider spatial coherence over a pyramid of scales.

As well as a measure, it may be possible to use our method directly to minimise the loss of a colourisation network to improve spatial coherency in colourisation. 
	
Our work is a full-reference performance measure. Following the assumptions of \cite{LevinColorizationOptimization2004:CUO:1015706.1015780}, \cite{SapiroInpaintingColors1530151} and \cite{YatzivChrominanceBlending1621234}, thought could be given to a spatial comparison between the L-channel prior and the predicted hue, thus enabling a no-reference performance measure.

\section{Conclusions}

We showed that widely-used pixel-difference measures of colourisation plausibility such as MSE and PSNR are misleading.  The hue channel is particularly problematic for these measures, as plausible colourisations can be drawn from a distribution of hues, but converting to other colour spaces propagates the hue problems to other channels. Using an $L^*h_{ab}^{*}c_{ab}^{*}$ formulation of CIEL*a*b*, the problems can be to compartmentalised into the hue channel only. We introduced a measure called SHSM for the hue channel, which we show measures the spatial coherence of hue-channel predictions in colourisation. We presented findings that current state-of-the-art CNN based colourisation methods exhibit high variance in spatial coherence. The SHSM approach will prove useful in progressing the field of colourisation, allowing researchers to assess the spatial coherence of their results both quantitatively and qualitatively.

\bibliographystyle{apalike}

\bibliography{refs.bib}
\section*{Supplementary Material}

Code for this paper can be found at https://github.com/seanmullery/SHSM

\end{document}